\documentclass[11pt]{article}

\usepackage{graphicx}
\usepackage{xcolor} % blue change highlighting
\usepackage{svg}
\usepackage{amsmath}
\usepackage{amsthm}
\usepackage{amssymb}
\usepackage{fancyhdr}
\usepackage{enumerate}
\usepackage[breaklinks]{hyperref}
\usepackage{lscape}
\usepackage{multirow}
\usepackage{relsize}
\usepackage{setspace}
\usepackage{natbib}
\usepackage{fullpage}
\usepackage{caption}
\usepackage{authblk}
\usepackage{totcount}
\usepackage{array}
\usepackage{longtable}
\usepackage{subcaption} % For subcaptions
\usepackage[titletoc]{appendix}
\usepackage{tikz}
\usetikzlibrary{positioning,calc} 
\usepackage[top    = 0.75in,
bottom = 1.22in,
left   = 1in,
right  = 1in]{geometry}
\usepackage{booktabs}
\usepackage{pgfplots}
\usetikzlibrary{arrows.meta}
\usepackage{algorithm}
\usepackage{algpseudocode}
\usepackage{makecell}
\usepackage{rotating}
\usepackage{tabularx}
\usepackage{booktabs}
\usepackage{array}

\newcolumntype{L}[1]{>{\raggedright\arraybackslash}p{#1}}

\newcolumntype{C}[1]{>{\centering\arraybackslash}m{#1}}

\newtotcounter{citnum} %From the package documentation
\def\oldbibitem{} \let\oldbibitem=\bibitem
\def\bibitem{\stepcounter{citnum}\oldbibitem}

\begin{document}
\onehalfspacing

\title{Reliable Evaluation of AI Agents for Investment Research: A Framework for Performance and Validity Assessment}

% \author{GILDATA}
% \date{ \today }

% \vspace{4mm}

\setcounter{Maxaffil}{10}

\renewcommand\Authsep{, }
\renewcommand\Authand{, }
\renewcommand\Authands{, }
\renewcommand\Authfont{\normalsize}
\renewcommand\Affilfont{\small}
\setlength{\affilsep}{1em}

\author{Dongnuan Tian}
\author{Yan Yu}
\author{Yangfan Liu}
\author{Wenjun Tang}
\author{Junting Kan}
\author{Xiang Wang}
\author{Sicong Liu}
\author{Yangyang Zhang}
\author{Shuanghong Li\thanks{Corresponding author. Email: lish58318@gildata.com}}

\affil{Gildata GAOS Team\\
Shanghai Hundsun JuYuan Data Service Co., Ltd.}

\date{ }

\pagestyle{fancy}
\fancyhf{}
\rhead{\thepage}
\lhead{\footnotesize \textbf{Tian et al.: }\textit{Reliable Evaluation of AI Agents for Investment Research}}

\normalsize
\maketitle

\vspace{-6mm}
\hrule
\text{ }\\

\noindent \Large \textbf{Abstract}\\
\normalsize

\noindent We consider the problem of evaluating AI agents undertaking investment-research tasks, where agents interact with external information sources and computational tools to produce financial analyses and research reports. An important challenge is that the quality of an agent's final output does not necessarily reflect the quality of its execution process, while the reliability of the resulting assessment also depends on the validity of the evaluation procedure. Furthermore, investment-research tasks differ considerably in their requirements for numerical accuracy, information retrieval, financial analysis and supporting evidence, making it difficult to assess agent performance using a uniform set of criteria. We propose a domain-oriented evaluation framework that distinguishes outcome quality from execution quality and accommodates these differences through task-specific criteria, deterministic validators and independent large language model judges. We also introduce an evaluation-validity mechanism that examines the integrity of evaluation inputs, reference information, supporting evidence and grading outputs, with expert review used to resolve identified deficiencies and significant grading disagreements. Numerical results show that the proposed framework can distinguish differences in agent performance across evaluation dimensions, while identifying potential reliability issues in the evaluation process. The associated implementation and evaluation materials will be made available on GitHub. \\%\textcolor{red}{\url{https://github.com/USERNAME/REPOSITORY}.}\\

\noindent \textit{Keywords: }
AI agent, investment research, performance assessment, evaluation validity\\

\hrule

\newpage

\section{Introduction}\label{sec:introduction}

Large language models (LLMs) have increasingly been used to develop AI agents capable of interacting with external tools and information sources to undertake complex tasks (see, e.g., \cite{yao2022react,wu2024autogen,wang2024survey,chowa2026language}). In contrast to conventional LLM applications in which a response is generated from a given input, tool-augmented agents can obtain additional information, perform intermediate computations, and determine subsequent actions according to the results of previous interactions. These capabilities are particularly relevant to investment research, where the preparation of financial analyses and research reports may involve information retrieval, quantitative calculations, and the interpretation of evidence from multiple sources \citep{zhou2024finrobot}. The resulting research output is therefore associated with an execution process comprising a sequence of decisions and tool interactions. Evaluating such an agent raises the question of how the quality of its final output and the quality of its execution process should be assessed, and whether the resulting assessments provide a reliable basis for comparing different agents.

A substantial body of research has investigated the evaluation of LLMs and AI agents, with methods developed for different task settings and assessment requirements \citep{chang2024survey,yehudai2026survey}. Conventional benchmark-based approaches typically evaluate task completion or the correctness of generated outputs against predefined criteria. However, these approaches may provide limited information about the intermediate actions undertaken by an agent or the quality of outputs for which a unique reference answer is unavailable. To address different aspects of these problems, researchers have considered execution-based evaluation and LLM-based assessment. For example, \citet{zhuge2024agent} proposed Agent-as-a-Judge, in which an evaluating agent examines task requirements and supporting artefacts through an interactive assessment process. \citet{es2024ragas} developed RAGAS to evaluate retrieval-augmented generation through criteria concerning the retrieved context and generated responses. Other studies have examined the consistency and reliability of LLM-based evaluators \citep{gu2026survey,tan2025judgebench}. These methods provide useful foundations for assessing agents, particularly when the assessment involves both directly verifiable requirements and qualitative judgements supported by external evidence.

In the financial domain, earlier benchmarks concentrated primarily on particular capabilities of language models, including numerical reasoning, financial knowledge, and question answering (\cite{lee2025large}). \citet{chen2021finqa} introduced FinQA to evaluate numerical reasoning over financial reports, while \citet{islam2023financebench} developed FinanceBench to examine the ability of LLMs to answer financial questions using supporting evidence. \citet{xie2024finben} subsequently proposed FinBen, which covers a broader collection of financial tasks. These studies establish useful approaches for evaluating financial reasoning through structured questions, reference information, and task-specific assessment criteria. Nevertheless, the evaluation of an investment-research agent may involve more than determining whether an individual financial question has been answered correctly. In particular, research tasks may require the agent to retrieve and combine information, develop analytical arguments, and produce outputs whose quality cannot be determined solely by comparison with a predefined answer.

Recent studies have thus considered more complex financial-research and investment-management settings. \citet{sun2025finresearchbench} proposed FinResearchBench, a logic-tree-based Agent-as-a-Judge framework for evaluating financial research reports. Their approach extracts the structure of arguments and supporting evidence from a report and assesses different aspects of research quality using both rule-based measures and LLM judgements. This provides a methodology for examining the analytical content of financial research beyond the correctness of individual answers. \citet{hu2026finsearchcomp} developed FinSearchComp to evaluate financial information retrieval and reasoning in settings involving complex searches and time-sensitive financial information. More recently, \citet{bek2026finskillbench} introduced FinSkillBench to assess agents undertaking investment-management tasks, incorporating point-in-time data, task-specific verifiers, and execution trajectories. These contributions are particularly relevant to the present research because they provide complementary approaches to evaluating analytical outputs, financial information retrieval, and the use of domain-specific tools. They also establish the importance of specifying the information and verification procedures associated with each financial task, for which the framework proposed in this paper (see Section~\ref{sec:evaluation_framework}) is a natural fit.

The development of these evaluation methods raises several methodological questions when assessing agents across different investment-research tasks. First, the requirements of financial tasks differ in the extent to which their outcomes can be verified objectively (\cite{sun2025finresearchbench}). Numerical results and explicitly specified requirements may be assessed through deterministic calculations or reference-based checks, whereas the quality of an analytical argument and its supporting evidence may require contextual judgement. The use of different grading mechanisms is already supported by existing research, including approaches that combine rule-based assessment and LLM evaluation \citep{sun2025finresearchbench,anthropic2026evals}. The relevant question is therefore not whether one grading mechanism is generally preferable to another, but how the mechanism should be selected and applied according to the requirements of a particular criterion. In addition, the correctness of financial information must be considered in relation to its reporting period and availability at the time specified by the task \citep{hu2026finsearchcomp,bek2026finskillbench}. Consequently, a common evaluation procedure must accommodate different forms of verification without treating all financial-research criteria as equivalent.

Second, the assessment of an agent's final output does not necessarily characterise its execution process. Existing agent-evaluation studies have demonstrated the relevance of intermediate actions and execution trajectories for understanding agent behaviour \citep{ma2024agentboard, zhuge2024agent,yehudai2026survey,bek2026finskillbench}. In investment research, the execution trace may provide evidence about information retrieval, tool selection, intermediate calculations, and the sequence in which research activities are undertaken. Such evidence is relevant to evaluating the appropriateness and efficiency of the execution process, but it does not replace an assessment of the final research output. Conversely, the correctness of the final output does not establish whether the agent has followed an appropriate execution process. We therefore distinguish outcome quality and execution quality as two components of agent performance. This distinction allows the assessment to consider both the research produced by an agent and the process through which it is obtained, while recognising that the criteria and evidence required for the two components may differ.

A third question concerns the reliability of the evaluation procedure used to obtain the performance scores. Previous research has examined limitations of LLM-based evaluation, including variation in judgements and the consistency of grading decisions \citep{gu2026survey,tan2025judgebench,stureborg2024large}. However, the reliability of an assessment also depends on whether the correct task requirements, reference information, and execution evidence have been supplied to the evaluator. An assessment may produce a numerical score even when the reference information does not correspond to the task, the available execution trace is incomplete, or the grading output does not satisfy the prescribed assessment requirements. These problems differ from errors made by the agent being evaluated: they concern whether the evaluation has been conducted using the appropriate information and procedures. Thus, the assessment of agent performance and the verification of the assessment procedure should be treated as separate questions. In particular, a low performance score should be distinguished from a score whose interpretation is affected by deficiencies in the evaluation process.

To address these methodological challenges, we make the following contributions:

\begin{itemize}
    \item We propose a domain-oriented framework for assessing the outcome and execution quality of AI agents undertaking investment-research tasks. The framework accommodates different task requirements and forms of evidence by specifying assessment criteria and selecting the corresponding grading mechanisms.

    \item We develop an evaluation-validity mechanism that verifies the association of evaluation inputs, the integrity of reference information and execution evidence, and the consistency of grading outputs. This enables deficiencies in the evaluation procedure to be distinguished from deficiencies in agent performance.

    \item We construct an investment-research evaluation testbed with task-specific criteria, reference information, and verification procedures, and conduct computational experiments to examine agent performance, evaluation-validity issues, and their implications for comparisons across tasks and agents\\
\end{itemize}

The remainder of the paper is organised as follows. Section~\ref{sec:motivation_problem} introduces the motivating example and problem formulation. Section~\ref{sec:evaluation_framework} presents the proposed evaluation framework. Section~\ref{sec:numerical_results} reports the numerical results, and Section~\ref{sec:conclusions} concludes the paper. The appendices provide a review of related work (Appendix~\ref{sec:related_work}), additional methodological details and supplementary results.\\

\section{Motivation and problem formulation}
\label{sec:motivation_problem}

\subsection{Motivation}
\label{subsec:motivation}

\begin{figure}[htbp]
    \centering
    \includegraphics[width=\linewidth]
        {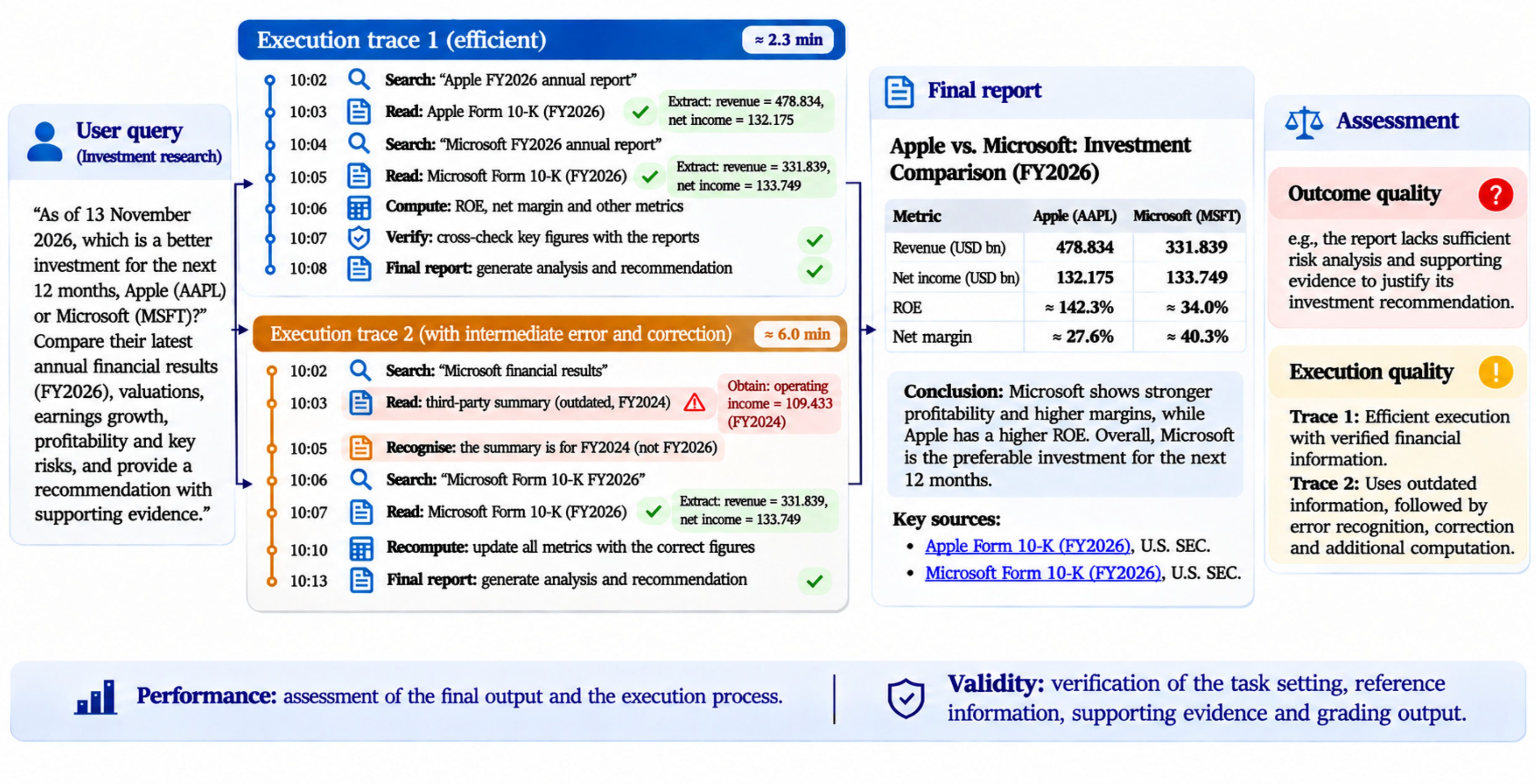}
    \caption{Illustrative evaluation of an investment-research task. Two execution traces produce the same final report through different processes. Although the report contains financial comparisons, its investment recommendation lacks sufficient risk analysis and supporting evidence. The example illustrates the distinction between outcome quality and execution quality, and the need to verify the validity of their assessments.}
    \label{fig:motivation_example}
\end{figure}

Consider an agent asked to compare the financial performance of two companies at a specified date. The resulting report provides evidence about the accuracy of the reported figures and the consistency of the financial analysis. However, these properties do not establish how the agent obtained the information or whether its intermediate actions were appropriate. Figure~\ref{fig:motivation_example} illustrates two possible executions that produce the same final result but differ in their execution traces. An assessment based only on the final report would not distinguish these processes, while an assessment based only on the trace would not establish whether the report adequately satisfies the task requirements. Moreover, the interpretation of either assessment depends on whether the required reference information and execution evidence are available and correspond to the specified task. We therefore distinguish the assessment of agent performance from the verification of the procedure used to obtain that assessment.

The example motivates two outputs of the evaluation procedure, namely a performance assessment and an evaluation-validity indicator. We formalise these outputs below and describe the corresponding verification procedure in Section~\ref{sec:evaluation_framework}.

\subsection{Problem formulation}
\label{subsec:problem_formulation}

Let $\mathcal{T}$ denote a set of investment-research tasks and $\mathcal{A}$ a set of AI agents. Each task $t\in\mathcal{T}$ is specified by its requirements and the information available at the time of execution. These specifications determine both the criteria against which the agent is assessed and the reference information that can be used for verification. When agent $a\in\mathcal{A}$ undertakes task $t$, it produces a final output $y_{a,t}$ and an execution trace $\tau_{a,t}$ recording its intermediate actions, tool interactions, and observations. We represent the resulting execution by
\begin{equation}
    x_{a,t}=(y_{a,t},\tau_{a,t}).
    \label{eq:agent_execution}
\end{equation}
For each task, let $c_t$ denote the assessment criteria and $r_t$ the corresponding reference information. The latter may include reference values, supporting documents, and their relevant temporal context, as required by the task. We define the evaluation record for agent $a$ undertaking task $t$ as
\begin{equation}
    z_{a,t}=(t,x_{a,t},c_t,r_t).
    \label{eq:evaluation_record}
\end{equation}

Agent performance is assessed from two perspectives. Outcome quality concerns the extent to which the final output satisfies the task requirements, whereas execution quality concerns the appropriateness and efficiency of the process through which that output is obtained. Let $f^{\mathrm{out}}$ and $f^{\mathrm{exe}}$ denote the corresponding assessment functions. The performance assessment is represented by
\begin{equation}
    \mathbf{s}_{a,t}
    =
    \left(
        \mathbf{s}^{\mathrm{out}}_{a,t},
        \mathbf{s}^{\mathrm{exe}}_{a,t}
    \right)
    =
    \left(
        f^{\mathrm{out}}(z_{a,t}),
        f^{\mathrm{exe}}(z_{a,t})
    \right).
    \label{eq:performance_assessment}
\end{equation}
Here, $\mathbf{s}^{\mathrm{out}}_{a,t}$ and $\mathbf{s}^{\mathrm{exe}}_{a,t}$ denote the dimension-specific scores for outcome and execution quality, respectively, with their dimensions and grading rules determined by the task-specific criteria $c_t$.

The performance scores alone do not establish whether the assessment procedure has been applied appropriately. We therefore introduce a separate validity function $g$ to examine the integrity of the evaluation inputs, supporting evidence, and grading procedure. For a completed assessment, let
\begin{equation}
    v_{a,t}=g(z_{a,t},\mathbf{s}_{a,t}),
    \qquad v_{a,t}\in\{0,1\},
    \label{eq:evaluation_validity}
\end{equation}
where $v_{a,t}=1$ indicates that the prescribed validity requirements are satisfied and $v_{a,t}=0$ indicates that at least one requirement is not satisfied. The indicator describes the validity of the assessment and should not be interpreted as a measure of agent performance.

The evaluation problem is to specify a procedure $\mathcal{E}$ that maps each evaluation record to its performance assessment and associated validity indicator:

\begin{equation}
    \mathcal{E}:z_{a,t}
    \longmapsto
    \left(\mathbf{s}_{a,t},v_{a,t}\right).
    \label{eq:evaluation_problem}
\end{equation}
The mapping in Equation~\eqref{eq:evaluation_problem} describes a completed assessment. If a preliminary integrity check fails, scoring is deferred and the corresponding record is not assigned a completed performance assessment until the deficiency has been examined. The validity indicator is therefore not a substitute for a score that has been withheld. The sequence of preliminary verification, performance assessment, and subsequent grading verification is described in the next section.\\

\section{Evaluation framework}
\label{sec:evaluation_framework}

\begin{figure}[htbp]
    \centering
    \includegraphics[width=\linewidth]
        {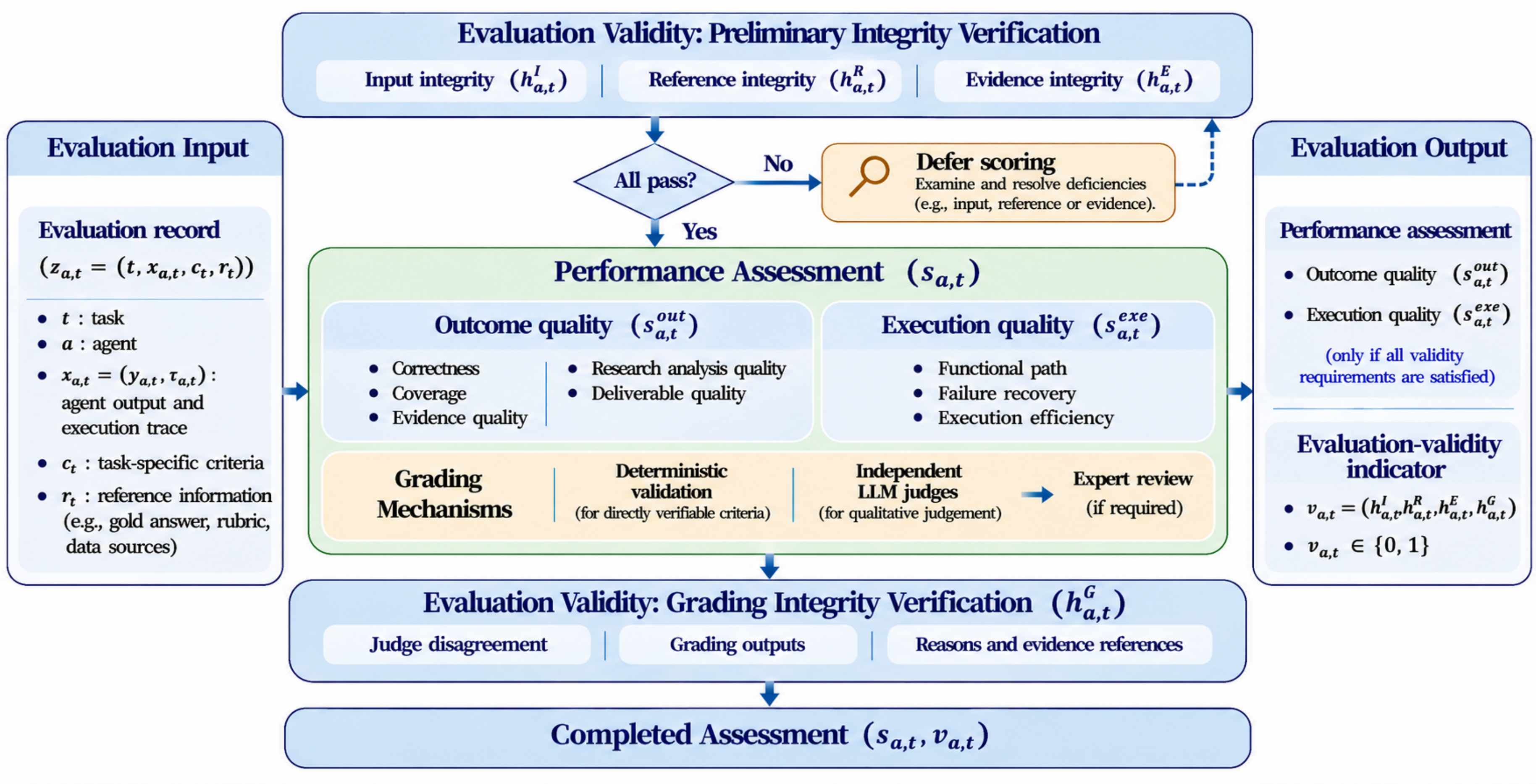}
    \caption{The proposed evaluation framework for investment-research agents. Each evaluation record is first examined for input, reference, and evidence integrity. Following these preliminary checks, outcome and execution quality are assessed using the applicable grading mechanisms. The resulting performance scores are then checked for grading integrity before the evaluation result is accepted. Scoring is deferred if a preliminary integrity requirement is not satisfied.}

\label{fig:evaluation_framework}
\end{figure}

We now describe the evaluation framework used to obtain the performance assessment and validity indicator defined in Section~\ref{subsec:problem_formulation}. The principal steps are illustrated in Figure~\ref{fig:evaluation_framework}.

The performance assessment applies the functions defined in Equation~\eqref{eq:performance_assessment} to the final output and execution trace contained in each evaluation record. Table~\ref{tab:performance_assessment} summarises the assessment focus and supporting evidence for outcome and execution quality. The applicable dimensions and scoring rules are specified by the task-specific criteria $c_t$, allowing different investment-research tasks to be assessed according to their respective requirements.

\begin{table}[htbp]
\centering
\caption{Assessment focus and supporting evidence for outcome and execution quality.}
\label{tab:performance_assessment}
\small
\renewcommand{\arraystretch}{1.25}
\begin{tabularx}{\linewidth}{
    >{\raggedright\arraybackslash}p{25mm}
    >{\raggedright\arraybackslash}X
    >{\raggedright\arraybackslash}p{34mm}
}
\toprule
\textbf{Component}
&
\textbf{Assessment focus}
&
\textbf{Evidence}
\\
\midrule
Outcome quality
&
Task completion, factual accuracy,
analytical reasoning, supporting evidence,
and output requirements.
&
Final output and reference information.
\\
Execution quality
&
Planning, tool usage, and the
appropriateness of the execution path.
&
Execution trace and tool interactions.
\\
\bottomrule
\end{tabularx}
\end{table}

For each assessment criterion, we select a grading mechanism according to the information needed to determine whether the criterion is satisfied. Deterministic validation is used where the result can be checked against an explicit rule or reference value, whereas criteria requiring interpretation of analytical arguments or supporting evidence are assessed by two independent, rubric-guided LLM judges. The judges' scores and supporting reasons are compared, and substantial disagreements are referred for expert review. Following the assessment of the applicable criteria and the resolution of identified disagreements, the resulting dimension-specific scores are consolidated according to the prescribed scoring rules to obtain the performance vector $\mathbf{s}_{a,t}$ defined in Equation~\eqref{eq:performance_assessment}.

The validity of the assessment also depends on whether the correct inputs and supporting evidence have been supplied and the grading procedure has been applied as specified. We therefore examine four integrity requirements: input integrity, reference integrity, evidence integrity, and grading integrity. Table~\ref{tab:validity_checks} specifies what is checked in each case and the conditions under which a check fails. The first three requirements are verified before scoring because they determine whether the assessment can proceed using the available record. Grading integrity is examined after scoring, when the grades, reasons, and evidence references produced by the evaluators are available.

\begin{table}[htbp]
\centering
\caption{Integrity requirements used to verify evaluation validity. The first three checks precede scoring; grading integrity is verified after the grading outputs have been produced.} 
\label{tab:validity_checks}
\small
\renewcommand{\arraystretch}{1.2}
\begin{tabularx}{\linewidth}{
    >{\raggedright\arraybackslash}p{25mm}
    >{\raggedright\arraybackslash}X
    >{\raggedright\arraybackslash}X
}
\toprule
Check &
Verification requirement &
Failure condition \\
\midrule
Input integrity &
Correct association of the task,
agent execution, assessment criteria,
and reference information. &
Missing or mismatched evaluation inputs. \\
Reference integrity &
Provenance, correctness, and temporal
consistency of reference information. &
Invalid sources, reporting periods,
or reference information. \\
Evidence integrity &
Availability and consistency of the
final output and execution trace required
for assessment. &
Missing or inconsistent assessment evidence. \\
Grading integrity &
Compliance with scoring rules,
completeness of grading outputs, and
consistency between scores, reasons,
and supporting evidence. &
Invalid grading outputs or unresolved
discrepancies. \\
\bottomrule
\end{tabularx}
\end{table}

For each evaluation record $z_{a,t}$, let

\begin{equation}
    \mathbf{h}_{a,t}
    =
    \left(
        h^{\mathrm I}_{a,t},
        h^{\mathrm R}_{a,t},
        h^{\mathrm E}_{a,t},
        h^{\mathrm G}_{a,t}
    \right)
    \in\{0,1\}^{4},
    \label{eq:validity_vector}
\end{equation}
denote the validity vector, where each component indicates whether the corresponding integrity requirement is satisfied. For a completed assessment, the overall validity indicator is given by

\begin{equation}
    v_{a,t}
    =
    \prod_{k\in\{\mathrm I,\mathrm R,
                   \mathrm E,\mathrm G\}}
    h^k_{a,t}.
    \label{eq:validity_aggregation}
\end{equation}
Thus, $v_{a,t}=1$ if and only if all four integrity requirements are satisfied, while the validity vector $\mathbf{h}_{a,t}$ identifies the outcome of each individual check. If a preliminary integrity check fails, scoring is deferred until the identified deficiency has been examined.\\

\section{Numerical results}\label{sec:numerical_results}

In this section, we report the results of an experimental evaluation of three AI agents, namely Codex, DSH and OpenClaw, on 62 investment-research tasks, resulting in 186 agent--task records. The task set consists of 14 factual retrieval and numerical accuracy tasks (T1), 15 multi-step financial analysis and comparison tasks (T2), 27 time-sensitive information and supporting evidence tasks (T3), and six open-ended research and interpretation tasks (T4). The distribution of tasks and corresponding records is provided in Appendix~\ref{app:nr_supplementary}. Agent performance is assessed using eight evaluation dimensions, comprising five outcome-quality dimensions (correctness, coverage, evidence quality, research analysis quality and deliverable quality) and three execution-quality dimensions (functional path, failure recovery and execution efficiency). The applicable dimensions and their weights are determined by the requirements of each task. When a dimension is not assessable, it is excluded from the calculation, and the weights of the remaining assessable dimensions are renormalised. The scoring perspectives and detailed calculations are provided in Appendices~\ref{app:nr_perspectives} and~\ref{app:nr_calculations}.

\subsection{Performance of the agents}\label{subsec:agent_performance_results}

Table~\ref{tab:nr_overall} presents the mean outcome-quality and execution-quality scores of the three agents, together with their raw and final overall scores under P0 (see Appendix~\ref{app:nr_calculations}). Codex achieves the highest mean scores for both outcome quality (82.07) and execution quality (86.05), followed by DSH (78.41 and 81.90) and OpenClaw (76.00 and 81.87). Although DSH and OpenClaw have almost identical execution-quality scores, DSH achieves a higher outcome-quality score. Expert-reviewed adjustments reduce the final overall scores relative to the raw scores for all three agents, with the largest difference observed for DSH.

\begin{table}[htbp]\centering\small
\caption{Overall performance of the three agents across 62 tasks (0--100 scale). Final overall scores incorporate expert-reviewed adjustments to raw scores.}\label{tab:nr_overall}
\begin{tabular}{lrrrrr}\toprule
Agent & Scored tasks & Outcome & Execution & Raw overall & Final overall\\\midrule
Codex & 62 & 82.07 & 86.05 & 83.04 & 82.94\\
DSH & 62 & 78.41 & 81.90 & 79.29 & 77.75\\
OpenClaw & 62 & 76.00 & 81.87 & 77.46 & 76.99\\\bottomrule\end{tabular}
\end{table}

Next, we examine the sensitivity of the comparison to the dimensions included in the assessment. Table~\ref{tab:nr_sensitivity} reports the percentage improvements of Codex and DSH relative to OpenClaw under two common-scope perspectives. Under P1, the three agents are compared using the same jointly assessable dimensions for each task. Under P2, execution efficiency is excluded and the remaining weights are renormalised. Codex outperforms OpenClaw by 7.20\% on average under P1, compared with 2.36\% for DSH. The corresponding improvements increase to 8.19\% and 3.19\% under P2. Thus, excluding execution efficiency increases the mean relative improvement of both agents, suggesting that this dimension contributes favourably to OpenClaw's performance in the original comparison. However, DSH remains below OpenClaw at the 10th percentile under P2, indicating that the improvement is not uniform across the score distribution. The percentile comparisons are calculated from corresponding percentiles of the agents' task-level scores, rather than from percentiles of task-by-task improvement ratios (see Appendix~\ref{app:nr_calculations}).

\begin{table}[htbp]\centering\small
\caption{Percentage improvements relative to OpenClaw under common-scope comparisons P1 and P2 (62 tasks). P2 excludes execution efficiency and renormalises the remaining weights.}\label{tab:nr_sensitivity}
\begin{tabular}{llrrrrrr}\toprule
Perspective & Agent & Mean & 10th & 25th & 50th & 75th & 90th\\\midrule
P1 & Codex & 7.20 & 0.00 & 11.14 & 9.42 & 1.91 & 3.59\\
   & DSH & 2.36 & -4.03 & -1.14 & 9.03 & 0.34 & 0.44\\\addlinespace
P2 & Codex & 8.19 & 3.94 & 11.29 & 11.07 & 3.26 & 3.59\\
   & DSH & 3.19 & -1.73 & 0.58 & 10.29 & 0.90 & 0.11\\\bottomrule
\end{tabular}
\end{table}

To investigate further, Tables~\ref{tab:nr_dims} and~\ref{tab:nr_types} present the results by evaluation dimension and task type, respectively. Table~\ref{tab:nr_dims} shows that Codex achieves the highest mean score in six of the eight dimensions, while DSH performs best in coverage and OpenClaw in execution efficiency. In particular, OpenClaw obtains a mean execution-efficiency score of 4.48 out of 5, compared with 4.14 for Codex and 4.06 for DSH. This difference is consistent with the sensitivity analysis in Table~\ref{tab:nr_sensitivity}, where excluding execution efficiency increases the relative improvements of the other two agents. Table~\ref{tab:nr_types} shows that Codex achieves the highest mean outcome-quality and execution-quality scores across all four task types. For all three agents, the mean outcome-quality score is lower for T2 than for T3. The results therefore indicate that relative performance depends on the dimension considered, while the absolute scores also vary with the requirements of different task types.

\begin{table}[htbp]\centering\small
\caption{Mean dimension scores (0--5 scale) on commonly assessable tasks.}\label{tab:nr_dims}
\begin{tabular}{p{18mm}p{52mm}rrrr}\toprule
Component & Dimension & Tasks & Codex & DSH & OpenClaw\\\midrule
Outcome & Correctness & 50 & 4.02 & 3.82 & 3.85\\
 & Coverage & 62 & 4.08 & 4.10 & 3.73\\
 & Evidence quality & 61 & 3.99 & 3.71 & 3.61\\
 & Research analysis quality & 47 & 4.26 & 4.16 & 4.00\\
 & Deliverable quality & 62 & 4.58 & 4.04 & 4.25\\\midrule
Execution & Functional path & 62 & 4.44 & 4.20 & 4.13\\
 & Failure recovery & 62 & 4.21 & 3.94 & 3.61\\
 & Execution efficiency & 62 & 4.14 & 4.06 & 4.48\\\bottomrule\end{tabular}
\end{table}

\begin{table}[htbp]\centering\scriptsize
\caption{Mean outcome-quality and execution-quality scores by task type (0--100 scale) on commonly assessable tasks.}\label{tab:nr_types}
\begin{tabular}{p{42mm}lrrrr}\toprule
Task type & Component & Tasks & Codex & DSH & OpenClaw\\\midrule
T1: Factual retrieval & Outcome & 14 & 82.81 & 79.64 & 72.50\\
 & Execution & & 86.33 & 84.92 & 82.85\\\addlinespace
T2: Multi-step financial analysis & Outcome & 15 & 73.25 & 67.59 & 68.21\\
 & Execution & & 83.31 & 79.14 & 74.51\\\addlinespace
T3: Time-sensitive research & Outcome & 27 & 86.86 & 85.51 & 81.58\\
 & Execution & & 86.73 & 82.53 & 84.19\\\addlinespace
T4: Open-ended research & Outcome & 6 & 80.86 & 70.68 & 78.58\\
 & Execution & & 89.18 & 78.89 & 87.52\\\bottomrule\end{tabular}
\end{table}

\subsection{Evaluation validity and grading consistency}\label{subsec:validity_results}

All 186 agent--task records contained the evidence required for assessment and were ultimately cleared for scoring, giving an assessment availability rate of 100\% (see Appendix~\ref{app:nr_supplementary}). Preliminary integrity checks were conducted before scoring, while grading integrity was verified after scoring. Table~\ref{tab:nr_integrity} presents the integrity failures identified during verification. No input-integrity failures were identified, while failures in the remaining three integrity checks were identified in 18.8\%, 15.1\% and 48.4\% of the records, respectively. All identified failures were subsequently resolved before the corresponding assessments were accepted. As multiple integrity requirements may fail for the same record, the failure counts cannot be added to determine the total number of affected records.

\begin{table}[htbp]\centering\small
\caption{Integrity failures across 186 agent--task records.}\label{tab:nr_integrity}
\begin{tabular}{p{69mm}rr}\toprule
Requirement & Failed runs & Failure rate (\%)\\\midrule
Input integrity & 0 & 0.0\\
Reference integrity & 35 & 18.8\\
Evidence integrity & 28 & 15.1\\
Grading integrity & 90 & 48.4\\\bottomrule\end{tabular}
\end{table}

The consistency of the independent LLM judges is also investigated using 484 dimension-level assessments with two eligible judgements. Of these, 236 (48.8\%) were referred for further examination, while the remaining 248 (51.2\%) did not require referral. Among the 236 referred assessments, 201 required review and 35 had proposed corrections. All referred assessments were subsequently resolved and accepted following expert review. The detailed counts and corresponding denominators are provided in Appendix~\ref{app:nr_supplementary}. The referral rate indicates that further examination was required for nearly half of the paired assessments. However, this does not necessarily imply that the corresponding judgements were incorrect.

\section{Conclusions}\label{sec:conclusions}

In this paper, we have developed an evaluation framework for AI agents undertaking investment-research tasks. The framework provides a basis for comparing agent performance across different assessment dimensions and task requirements, while examining whether the resulting scores satisfy the prescribed evaluation requirements. Our numerical results show that Codex achieves the highest overall performance among the three agents considered, although the relative performance of the agents varies across evaluation dimensions and task types. The experiments also identify deficiencies in evaluation integrity and substantial disagreements between independent judges, indicating the importance of verifying the evaluation procedure before accepting the resulting scores. Future research could extend the framework to a wider range of investment-research settings and investigate the sensitivity of the results to different grading criteria and evaluators.

% \section*{Acknowledgments.}

% References here (outcomment the appropriate case) 

% CASE 1: BiBTeX used to constantly update the references 
%   (while the paper is being written).
\small
\bibliographystyle{apalike} % outcomment this and next line in Case 1
\bibliography{my_collection} % if more than one, comma separated

\newpage
\Large \textbf{Online Appendices}\\
\normalsize

\begin{appendices}

\section{Related work}
\label{sec:related_work}

The evaluation problem considered in this paper is closely related to four areas of research, namely, LLM-as-a-Judge, Agent-as-a-Judge, financial-agent evaluation, and evaluation reliability. We discuss the relevance of these approaches to the proposed framework below.\\

\noindent\textbf{LLM-as-a-Judge.}
LLM-based evaluators have been used to assess generated responses against criteria for which deterministic verification may be difficult. Existing approaches include rubric-guided assessment, reference-based evaluation, and methods that examine the relationship between generated responses and supporting information \citep{zheng2023judging,liu2023geval,kim2024prometheus,es2024ragas,saadfalcon2024ares,gu2026survey}. These methods are relevant to investment-research tasks because the quality of analytical arguments and supporting evidence may require contextual judgement. However, other requirements, such as numerical accuracy and compliance with explicitly specified conditions, can be assessed using deterministic rules. In our framework, the grading mechanism is therefore selected according to the requirements of each assessment criterion, rather than applying LLM-based evaluation uniformly. The reliability of the resulting judgements is examined separately from the selection of the grading mechanism.

\medskip
\noindent\textbf{Agent-as-a-Judge.}
The evaluation of AI agents requires consideration of the actions undertaken during task execution, which may not be fully reflected in the final output \citep{yehudai2026survey}. AgentBench and AgentBoard considered agent behaviour in interactive environments and the information provided by intermediate progress \citep{liu2024agentbench,ma2024agentboard}. \citet{zhuge2024agent} proposed Agent-as-a-Judge, in which an evaluating agent examines task requirements and supporting artefacts through a multi-step assessment process. In the financial domain, \citet{sun2025finresearchbench} proposed FinResearchBench, which extracts a logic tree from the generated research report to examine its analytical structure and supporting evidence. These studies demonstrate different ways of incorporating intermediate information into agent evaluation. In comparison, our framework specifies task-specific assessment criteria and verification requirements independently of the agent's generated output, and distinguishes evidence used to assess execution quality from evidence used to verify the evaluation procedure.

\medskip
\noindent\textbf{Financial-Agent Evaluation.}
Financial benchmarks have examined numerical reasoning, question answering, and broader financial capabilities using structured tasks and reference information \citep{chen2021finqa,islam2023financebench,xie2024finben}. More recent studies have extended evaluation to financial research reports, information retrieval, and investment-management tasks \citep{sun2025finresearchbench,bigeard2025financeagent,hu2026finsearchcomp,bek2026finskillbench,zhu2026finmcp}. FinResearchBench (\citealp{sun2025finresearchbench}) provides a method for examining analytical arguments and supporting evidence, while FinSearchComp (\citealp{hu2026finsearchcomp}) and FinSkillBench (\citealp{bek2026finskillbench}) consider requirements associated with time-sensitive information and task-specific verification. Research concerning risks associated with financial agents further highlights the importance of examining the conditions under which their outputs and behaviour are assessed \citep{chen2025standard}. These approaches address different aspects of financial-agent evaluation. Our framework considers outcome and execution quality within a common assessment procedure, while specifying the reference information, evidence, and verification requirements applicable to each task.

\medskip
\noindent\textbf{Evaluation Reliability.}
The reliability of model-based grading has been examined in relation to evaluator bias, agreement, consistency, and calibration \citep{zheng2023judging,wang2024fair,stureborg2024large,tan2025judgebench,gu2026survey,saadfalcon2024ares}. Related guidance has also considered the complementary roles of deterministic, model-based, and human assessment \citep{anthropic2026evals}. In financial applications, the correctness of reference information must additionally be considered in relation to reporting periods and information availability \citep{hu2026finsearchcomp,bek2026finskillbench}. These considerations concern different sources of uncertainty in an evaluation. In particular, consistency between grading judgements does not establish that the correct task requirements, reference information, and supporting evidence have been used. We therefore introduce separate integrity checks for these elements and for the resulting grading outputs. These checks establish compliance with prescribed evaluation requirements, rather than guaranteeing the correctness of every grading judgement.\\

\section{Evaluation perspectives for agent performance comparison}\label{app:nr_perspectives}

We consider three scoring perspectives to distinguish an agent's performance on its own valid assessment scope from comparisons based on a common set of tasks and dimensions. We use the agent and task sets $\mathcal A$ and $\mathcal T$ defined in Section~\ref{subsec:problem_formulation}. Let $V_a\subseteq\mathcal T$ be the set of tasks with valid assessments for agent $a\in\mathcal A$. For each $t\in V_a$, let $A_{a,t}$ be the set of dimensions that are applicable and assessable for the corresponding evaluation record $z_{a,t}$ (Equation~\eqref{eq:evaluation_record}). We define the common task set and the corresponding common dimension sets by
\begin{equation}\label{eq:nr_common_sets}
T^{\mathrm{com}}=\bigcap_{a\in\mathcal A}V_a,\qquad C_t=\bigcap_{a\in\mathcal A}A_{a,t},\quad t\in T^{\mathrm{com}}.
\end{equation}
Thus, $T^{\mathrm{com}}$ contains the tasks valid for all three agents, while $C_t$ contains the dimensions assessable for all three agents on task $t$. A task or dimension that does not satisfy the relevant validity requirements is excluded from the corresponding comparison rather than assigned a score of zero.

\paragraph{P0: Agent-specific absolute performance.}
Under P0, we calculate each agent's mean score using its valid task set $V_a$ and the corresponding assessable dimensions $A_{a,t}$. Since the eligible tasks and dimensions may differ between agents, this perspective describes performance under each agent's own assessment scope. It does not necessarily provide a comparison based on identical assessment conditions.

\paragraph{P1: Common-scope comparison.}
Under P1, we restrict the comparison to tasks in $T^{\mathrm{com}}$ and use the same dimension set $C_t$ and corresponding weights for all three agents on each task. Consequently, differences in the resulting scores are not attributable to differences in the included task sample or assessment dimensions. We use OpenClaw as the benchmark when calculating the percentage improvements reported in Table~\ref{tab:nr_sensitivity}.

\paragraph{P2: Common-scope comparison excluding execution efficiency.}
P2 uses the same common task set as P1 but excludes the execution-efficiency dimension from each $C_t$. We renormalise the weights over the remaining assessable dimensions. In particular, functional path and failure recovery remain included when they are jointly assessable. Comparing P1 with P2 therefore allows us to examine the effect of excluding execution efficiency, rather than comparing overall performance with outcome quality alone.

The three perspectives distinguish agent-specific performance estimates from comparisons based on identical assessment scopes. Their numerical definitions are given in Appendix~\ref{app:nr_calculations}.\\

\section{Calculation of performance and evaluation measures}\label{app:nr_calculations}

We describe the aggregation of the dimension-specific performance assessments $\mathbf{s}_{a,t}$ defined in Equation~\eqref{eq:performance_assessment}. In the experiments, $\mathcal T$ contains 62 tasks and $\mathcal A=\{\mathrm{Codex},\mathrm{DSH},\mathrm{OpenClaw}\}$. For each valid record $z_{a,t}$, let $s_{a,t,d}\in[0,5]$ denote the score of assessable dimension $d\in A_{a,t}$, where $s_{a,t,d}$ is a component of the outcome- or execution-quality score vector in $\mathbf{s}_{a,t}$. Let $w_{t,d}\geq0$ be the weight prescribed by the task-specific assessment criteria $c_t$ (Equation~\eqref{eq:evaluation_record}). The aggregate outcome-quality, execution-quality and overall scores use a 0--100 scale; multiplication by 20 converts a weighted score on the 0--5 scale to this scale.

\subsection{Task-level scores and expert-reviewed adjustments}

We first calculate the task-level overall score. Since the assessable dimensions may vary between agent--task records, we renormalise the task-specific weights over $A_{a,t}$, provided that their sum is positive:
\begin{equation}\label{eq:nr_weights}
\widetilde w_{a,t,d}=\frac{w_{t,d}}{\sum_{j\in A_{a,t}}w_{t,j}},\quad d\in A_{a,t},
\qquad
R_{a,t}=20\sum_{d\in A_{a,t}}\widetilde w_{a,t,d}s_{a,t,d}.
\end{equation}
Here, $R_{a,t}$ is the raw overall score of the evaluation record. An unassessable dimension is excluded from both the weighted sum and its normalisation. To calculate the outcome- and execution-quality scores separately, let $\mathcal D^{\mathrm{out}}$ and $\mathcal D^{\mathrm{exe}}$ denote their respective dimension sets, corresponding to the two components of $\mathbf{s}_{a,t}$ in Equation~\eqref{eq:performance_assessment}. For $c\in\{\mathrm{out},\mathrm{exe}\}$, define $A_{a,t}^{c}=A_{a,t}\cap\mathcal D^{c}$. The component scores and the corresponding sums of applicable weights are
\begin{equation}\label{eq:nr_components}
Q_{a,t}^{c}=20\frac{\sum_{d\in A_{a,t}^{c}}w_{t,d}s_{a,t,d}}{W_{a,t}^{c}},\qquad
W_{a,t}^{c}=\sum_{d\in A_{a,t}^{c}}w_{t,d},\quad c\in\{\mathrm{out},\mathrm{exe}\}.
\end{equation}
Thus, $Q_{a,t}^{\mathrm{out}}$ and $Q_{a,t}^{\mathrm{exe}}$ are the aggregated scores corresponding to the two performance components. A component score is defined only when its applicable weight $W_{a,t}^{c}$ is positive. When both components are defined, the overall score satisfies
\begin{equation}\label{eq:nr_component_overall}
R_{a,t}=\frac{W_{a,t}^{\mathrm{out}}Q_{a,t}^{\mathrm{out}}+W_{a,t}^{\mathrm{exe}}Q_{a,t}^{\mathrm{exe}}}{W_{a,t}^{\mathrm{out}}+W_{a,t}^{\mathrm{exe}}}.
\end{equation}
The overall score is therefore a weighted combination of the two components, rather than necessarily their arithmetic mean. For agent $a$, we denote its mean raw overall score by $\overline R_a$, the net expert-reviewed adjustment by $\Delta_a$, and the final overall score by $F_a$:
\begin{equation}\label{eq:nr_final}
\overline R_a=\frac{1}{|V_a|}\sum_{t\in V_a}R_{a,t},\qquad F_a=\overline R_a-\Delta_a.
\end{equation}
Here, $\Delta_a$ is the net difference between the raw and final mean scores following expert review. The reported data specify this adjustment only at the aggregate level; Equation~\eqref{eq:nr_final} does not imply a particular allocation of adjustments to individual tasks.

\subsection{Perspective-specific aggregation and percentage improvements}

To compare scores under different assessment scopes, we define a task-level score for any nonempty set $B$ of valid dimensions with positive total weight:
\begin{equation}\label{eq:nr_scope_score}
S_{a,t}(B)=20\frac{\sum_{d\in B}w_{t,d}s_{a,t,d}}{\sum_{d\in B}w_{t,d}},\qquad B\subseteq A_{a,t}.
\end{equation}
Let $d_{\mathrm{eff}}$ denote the execution-efficiency dimension. Using the task and dimension sets in Equation~\eqref{eq:nr_common_sets}, the mean scores under P0, P1 and P2 are
\begin{equation}\label{eq:nr_perspective_means}
\begin{aligned}
\overline S_a^{P0}&=\frac{1}{|V_a|}\sum_{t\in V_a}S_{a,t}(A_{a,t}),\\
\overline S_a^{P1}&=\frac{1}{|T^{\mathrm{com}}|}\sum_{t\in T^{\mathrm{com}}}S_{a,t}(C_t),\\
\overline S_a^{P2}&=\frac{1}{|T^{\mathrm{com}}|}\sum_{t\in T^{\mathrm{com}}}S_{a,t}(C_t\setminus\{d_{\mathrm{eff}}\}).
\end{aligned}
\end{equation}
The P0 score in Equation~\eqref{eq:nr_perspective_means} equals $\overline R_a$ in Equation~\eqref{eq:nr_final}; the expert-reviewed adjustment is not included in the P1 and P2 comparisons. These expressions require a nonempty eligible task set and a positive total weight for the dimensions included in each task-level score. In the present experiments, all 62 tasks are included in the P2 comparison, since each task retains positively weighted dimensions after execution efficiency is excluded. More generally, a task with no remaining positively weighted dimensions would have to be excluded from this comparison.

We calculate the relative improvements with respect to OpenClaw. Let $S_{a,q}^{Pk}$ denote the $q$th percentile of the task-level scores under perspective $Pk$, where $k\in\{1,2\}$ and $q\in\{10,25,50,75,90\}$. For the comparison of means, we use $q=\mathrm{mean}$ and define $S_{a,\mathrm{mean}}^{Pk}=\overline S_a^{Pk}$. Taking $b=\mathrm{OpenClaw}$ as the benchmark, the percentage improvement is
\begin{equation}\label{eq:nr_improvement}
I_{a,q}^{Pk}=100\frac{S_{a,q}^{Pk}-S_{b,q}^{Pk}}{S_{b,q}^{Pk}},\qquad S_{b,q}^{Pk}>0.
\end{equation}
For each value of $q$, we compare the corresponding score summaries for the two agents. In particular, the percentile-based improvements are not percentiles of the task-by-task improvement ratios. The P1 and P2 comparisons use the task-level scores under their respective assessment scopes; the aggregate expert-reviewed adjustments to the final overall scores are treated separately.

\subsection{Assessment availability and review rates}

Finally, we calculate assessment availability, integrity failures and independent-judge review rates using their respective units of analysis. For the availability rate, the numerator counts records scored following clearance; for integrity failure rates, the numerator counts records that failed the specified check during verification, before the identified deficiencies were resolved. The rates are
\begin{equation}\label{eq:nr_rates}
\begin{aligned}
\text{Availability rate}&=100\frac{N_{\mathrm{scored}}}{N_{\mathrm{expected}}},\\
\text{Failure rate for check }k&=100\frac{N_{k,\mathrm{failed}}}{N_{\mathrm{records}}},\\
\text{Referral rate}&=100\frac{N_{\mathrm{referred}}}{N_{\mathrm{paired}}},\\
\text{Resolution rate}&=100\frac{N_{\mathrm{resolved}}}{N_{\mathrm{referred}}}.
\end{aligned}
\end{equation}
For $k\in\{\mathrm I,\mathrm R,\mathrm E,\mathrm G\}$, the check corresponds to a component of the validity vector $\mathbf h_{a,t}$ in Equation~\eqref{eq:validity_vector}. The failure counts describe deficiencies detected during verification, not the final value of $v_{a,t}$ after resolution (Equation~\eqref{eq:validity_aggregation}). Each integrity failure rate uses the 186 agent--task records as its denominator. The numbers of records failing input, reference, evidence and grading integrity are 0, 35, 28 and 90, respectively (Table~\ref{tab:nr_integrity}). Since one record may fail several requirements, the counts are not mutually exclusive. In contrast, the independent-judge referral rate is calculated over 484 eligible paired dimension-level assessments, of which 236 were referred and 248 were not. The resolution rate uses the 236 referred assessments as its denominator. These populations must be distinguished because a paired dimension-level judgement and an agent--task record represent different units of analysis.\\

\section{Supplementary results of the experimental evaluation}\label{app:nr_supplementary}

We provide additional information on the experimental task set, the availability of valid assessments and the review of independent-judge outputs. These results support the comparisons in Section~\ref{sec:numerical_results} and clarify the populations used to calculate the reported rates.

\subsection{Task distribution and assessment availability}

Table~\ref{tab:nr_design} shows the distribution of the 62 tasks across the four task types. Since each of the three agents undertook every task, the experiment contains 186 expected agent--task records. The largest group is T3, with 27 tasks and 81 records, while T4 contains six tasks and 18 records. Table~\ref{tab:nr_accept} reports the availability of assessment evidence and the completion of preliminary integrity checks. All 186 records contained the required evidence and were cleared for scoring following preliminary verification, giving an availability rate of 100\%. This does not imply that no deficiencies were identified during verification. Table~\ref{tab:nr_integrity} reports the failures detected during verification and subsequently resolved. Grading integrity was verified after scoring.

\begin{table}[htbp]\centering\small
\caption{Classification of the 62 evaluated investment-research tasks and corresponding agent--task records.}\label{tab:nr_design}
\begin{tabular}{lp{77mm}rr}\toprule
Type & Principal task requirement & Tasks & Expected records\\\midrule
T1 & Factual retrieval and numerical accuracy & 14 & 42\\
T2 & Multi-step financial analysis and comparison & 15 & 45\\
T3 & Time-sensitive information and supporting evidence & 27 & 81\\
T4 & Open-ended research and interpretation & 6 & 18\\\midrule
Total & & 62 & 186\\\bottomrule\end{tabular}
\end{table}

\begin{table}[htbp]\centering\small
\caption{Assessment availability and pre-scoring clearance for the 62-task evaluation.}\label{tab:nr_accept}
\begin{tabular}{lrrrrr}\toprule
Agent & Expected & Evidence & Pre-scoring pass & Scored & Availability (\%)\\\midrule
Codex & 62 & 62 & 62 & 62 & 100.0\\
DSH & 62 & 62 & 62 & 62 & 100.0\\
OpenClaw & 62 & 62 & 62 & 62 & 100.0\\\midrule
Total & 186 & 186 & 186 & 186 & 100.0\\\bottomrule\end{tabular}
\end{table}

\subsection{Independent-judge review outcomes}

Table~\ref{tab:nr_review} presents the outcomes of the independent-judge review procedure. Among 484 dimension-level assessments with two eligible judgements, 248 were accepted without referral and 236 were referred for further examination. The referred assessments comprise 201 cases requiring review and 35 cases with proposed corrections. All 236 referrals were subsequently resolved and accepted following expert review. Thus, the referral rate is 48.8\%, while the resolution rate among referred assessments is 100\%. These rates describe different stages of the evaluation procedure and should not be interpreted as measures of judgement accuracy. In particular, the aggregate data do not identify the causes of individual referrals or the number of incorrect initial judgements. Moreover, these counts concern dimension-level assessments, rather than the 186 agent--task records used to calculate the integrity-failure rates.

\begin{table}[htbp]\centering\small
\caption{Independent-judge disagreement referrals and review outcomes.}\label{tab:nr_review}
\begin{tabular}{p{69mm}p{37mm}rr}\toprule
Measure & Denominator & Count & Rate (\%)\\\midrule
Assessments with two eligible judgements & All eligible paired assessments & 484 & 100.0\\
Not referred & 484 paired assessments & 248 & 51.2\\
Referred for further examination & 484 paired assessments & 236 & 48.8\\
\quad Review required & 484 paired assessments & 201 & 41.5\\
\quad Correction proposed & 484 paired assessments & 35 & 7.2\\
Referrals resolved and accepted & 236 assessments & 236 & 100.0\\\bottomrule\end{tabular}
\end{table}

\end{appendices}

%%%%%%%%%%%%%%%%%
\end{document}